\documentclass[journal,10pt]{IEEEtran} 

\usepackage{makecell}

\usepackage{cite}

\usepackage{graphicx}

\ifCLASSINFOpdf
\else
\fi
\usepackage{amsmath}
\ifCLASSOPTIONcompsoc
 \usepackage[caption=false,font=normalsize,labelfont=sf,textfont=sf]{subfig}
\else
\usepackage[caption=false,font=footnotesize]{subfig}
\fi
\usepackage[font=footnotesize]{subfig}
\usepackage{url}

\usepackage[ruled,linesnumbered]{algorithm2e}

\usepackage{amsmath,amssymb}
\usepackage{xcolor,colortbl}

\newtheorem{theorem}{Theorem}[section]

\newtheorem{remark}[theorem]{Remark}

\newdimen\Lwidth
\begin{document}
\title{3D Orientation Field Transform}
\author{Wai-Tsun~Yeung,~\IEEEmembership{}
        Xiaohao~Cai,~\IEEEmembership{}
        Zizhen~Liang,~\IEEEmembership{}
        and~Byung-Ho~Kang,~\IEEEmembership{}
        \thanks{W-T. Yeung, Z. Liang, and B-H. Kang are with School of Life Sciences, The Chinese University of Hong Kong, Shatin, NT, Hong Kong (e-mail: bkang@cuhk.edu.hk).}
        \thanks{X. Cai is with School of Electronics and Computer Science, University of Southampton, University Road, Southampton, SO17 1BJ, UK (e-mail: x.cai@soton.ac.uk).}     
        }%
        \maketitle
\begin{abstract}
The two-dimensional (2D) orientation field transform has been proved to be effective at enhancing 2D contours and curves in images by means of top-down processing. It, however, has no counterpart in three-dimensional (3D) images due to the extremely complicated orientation in 3D compared to 2D. Practically and theoretically, the demand and interest in 3D can only be increasing. In this work, we modularise the concept and generalise it to 3D curves. Different modular combinations are found to enhance curves to different extents and with different sensitivity to the packing of the 3D curves. In 
principle, the proposed 3D orientation field transform can naturally tackle any dimensions. As a special case, it is also ideal for 2D images, owning simpler methodology compared to the previous 2D orientation field transform. 
The proposed method is demonstrated with several transmission electron microscopy tomograms ranging from 2D curve enhancement to, the more important and interesting, 3D ones. 
\end{abstract}

\begin{IEEEkeywords}
Orientation field transform, 3D, image segmentation, image denoising, electron tomography, curves.
\end{IEEEkeywords}

\section{Introduction}
\label{intro}
\begin{figure}[!tp]
\centering
\begin{tabular}{c}
\includegraphics[width=0.40\textwidth]{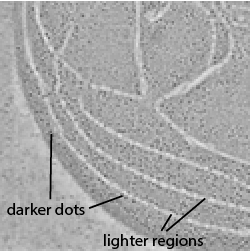} 
\end{tabular}
\caption{A 2D cross-section of a 3D electron tomogram (enlarged Fig. \ref{origimages} (a)). The darker dots and the lighter tubular regions are macromolecules and membrane-bound compartments, respectively. } 
\label{etexample}
\end{figure}

\IEEEPARstart The segmentation of transmission electron micrographs poses its own set of challenges, namely the low signal-to-noise ratio \cite{zanetti2009contrast} and the monotonicity of information, characterised by a single electromagnetic wave source and the difficulty of differential labelling \cite{adams2016multicolor}. When transmission electron microscopy is combined with computational tomography to produce three-dimensional (3D) images, it poses the additional problem of anisotropic resolution because of incomplete frequency information around the $z$-axis \cite{IEEEhowto:deng}. With biological samples comprised mostly of light atoms, imaging is achieved by fixing the sample and staining it with highly oxidising heavy metallic compounds. Such images are typically identifiable with curves denoting strands, lighter regions characterising membrane-bound compartments, and ubiquitous dots representing macromolecules, see Fig. \ref{etexample}. 

\begin{figure*}[!tp]
\centering
\begin{tabular}{ccccc}
\includegraphics[width=0.18\textwidth]{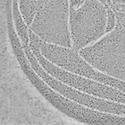} &
\includegraphics[width=0.18\textwidth]{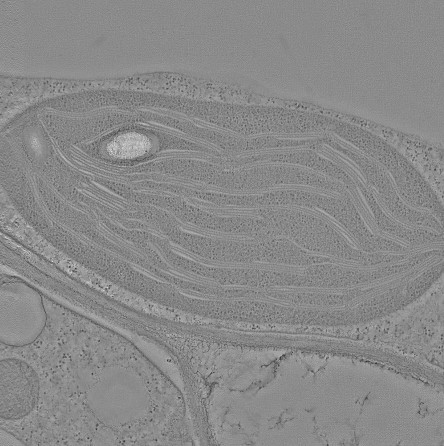} & 
\includegraphics[width=0.18\textwidth]{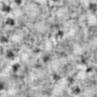} &
\includegraphics[width=0.18\textwidth]{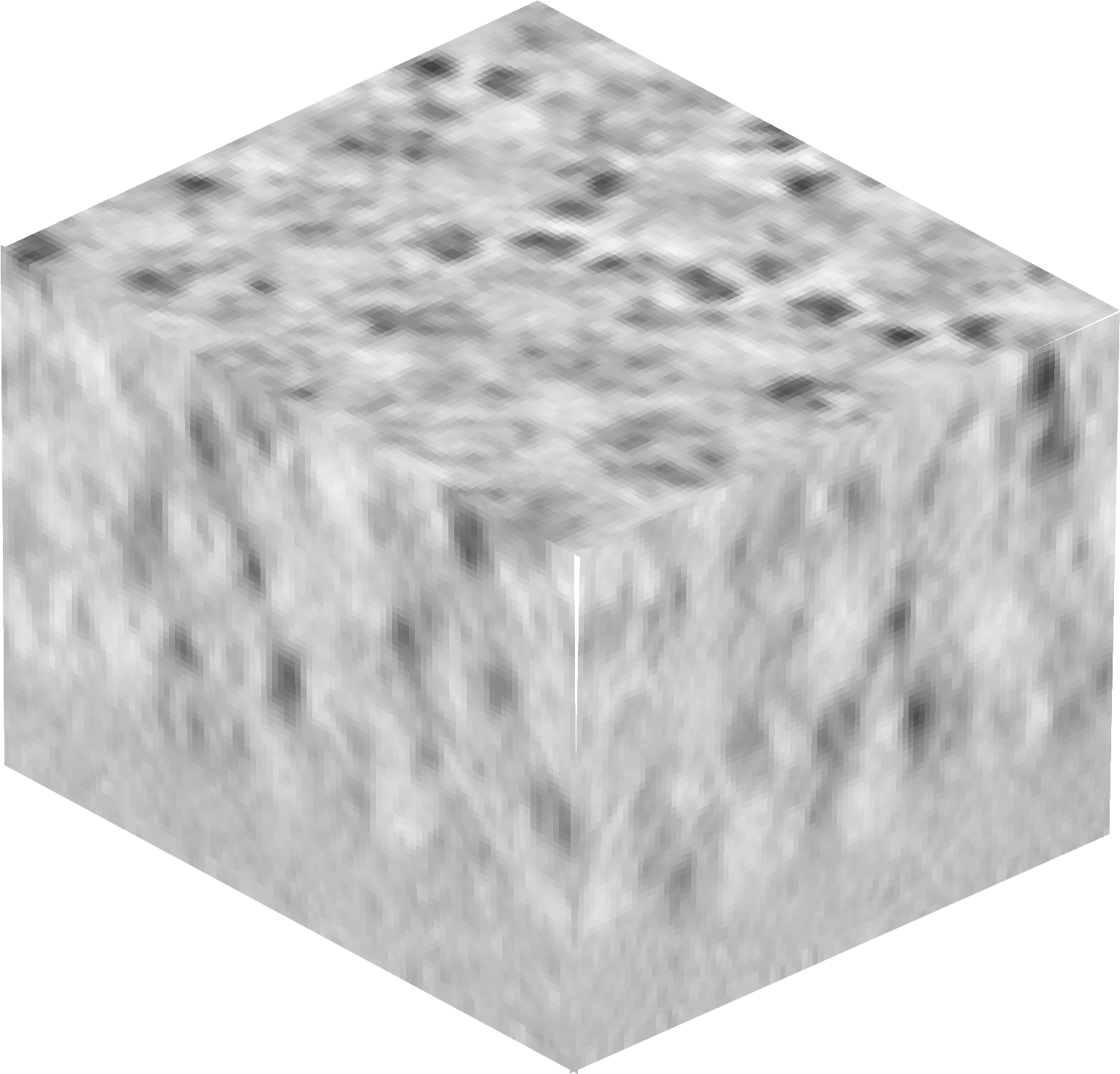} &
\includegraphics[width=0.18\textwidth]{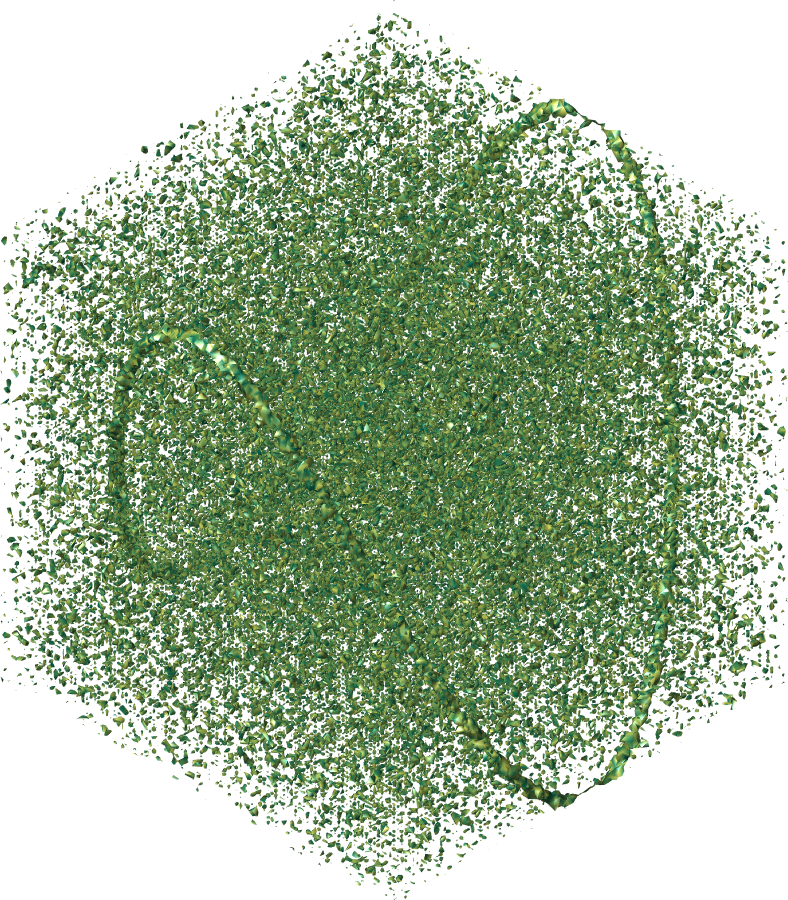} \\
(a) & (b) & (c) & (d) & (e)
\end{tabular}
\caption{Curves to be enhanced (and segmented) in the test images (refer to Section \ref{sec:experiment} for detailed description). Panels (a) and (b) are two types of 2D cross-sections of connected lipid membrane-bound compartments; panel (c) is a slice of a 3D liquid crystal data shown in (d); and panel (e) is a mesh of a 3D curve made for demonstration. The curve in panel (e) is a binary image created by hard thresholding a 3D noisy image which is generated by parametric equations $x(t)=\sin{t}$, $y(t)=\cos{t}$ and $z(t)=\cos{2t}$ with Gaussian noise.}
\label{origimages}
\end{figure*}

The ultimate aim of this paper is to enhance the curves in the 2D lamellae in plastids (see Fig. \ref{origimages} (a)--(b)) and the tubules in 3D lipid crystals (see Fig. \ref{origimages} (c)--(d)) so that the curve-like structures could be identified easily. The tubular architectures of the lipid crystals could be seen as scaffolds made of rods or curves in 3D space, see Fig. \ref{origimages} (d). 
Although the lamellar compartments are in fact curved sheets in 3D, their cross-sections could also be seen as curves in 2D images. In other words, lamellar compartments and tubules could be treated as curves in 2D and 3D images, respectively.
With the improving image technology in producing 3D images, there is an increasing need for curve enhancement methods that target 3D structures like the lipid crystal segmentation problem introduced above, which could facilitate for example the segmentation of curve-like structures.

In comparison to other types of images (e.g. medical imaging) in 2D/3D, there has yet to be tried-and-tested methods for reliable segmentation in electron tomography. As a result, it is still a common practice to do manual contouring to segment structures of interest \cite{frank2008electron,volkmann2010methods}.

Prevailing existing autosegmentation approaches include: 1) general noise-reduction techniques, most common of which are different variations of wavelet transform \cite{frank2008electron,huang2018exploring,stoschek1997denoising,volkmann2010methods,CCMS12,CCMS13}, nonlinear anisotropic diffusion or bilateral filtering \cite{bazan2009structure,frangakis1999nonlinear,frank2008electron,volkmann2010methods}; 2) direct segmentation techniques such as thresholding \cite{frank2008electron,CS13,CCZ13,CCSSZ19}, morphological operations \cite{frank2008electron}, region-based approaches utilising watershed transform \cite{frank2008electron,volkmann2002novel,volkmann2010methods}, and energy-based approaches in the manner of active contour \cite{frank2008electron,C15,tasel2016validated,volkmann2010methods}. 
Moreover, there lately have also been attempts at using machine-learning algorithms to improve the segmentation quality, e.g. \cite{khadangi2020net,staniewicz2015machine,yang2020subcellular}. As both electron tomography and machine learning are fairly new tools that have developed rapidly in the last decade, this would make a new frontier of research. 

The popular methods mentioned above however suffer from major shortcomings for the images that are being tested. For example, nonlinear anisotropic diffusion requires a relatively sharp contrast between objects in order to operate; watershed transform does not work with objects with a high genus number, which is a characteristic of the test images in this paper (Fig. \ref{origimages}); and wavelet transform and active contouring would require labour-intensive fine-tuning of ambiguous parameters. Recent developments of segmentation algorithms often rely on the integration of the aforementioned methods. As they are not mutually exclusive, mixing them is often a reliable technique at the expense of computational power and time.

In \cite{IEEEhowto:sandberg2007,IEEEhowto:sandberg2009}, 2D orientation field-based methods were proposed for vascular enhancement and segmentation. The advantage of this type of method is the capability of segmenting structures with a relatively high noise level, as long as the segmented objects in question bear semblance of a line or a curve, which fits the descriptions of the prolamellar body's tubules; moreover, it has only one single parameter with a clear physical meaning. This method however does not work with 3D lines or curves, which warrants the modification of the method that will be the focus of this paper.

The remainder of this paper is organised as follows. In Section \ref{sec:preliminaries} mathematical preliminaries are introduced. Then the related work, 2D orientation field transform, is presented in Section \ref{sec:relatedwork}. 
 In Section \ref{sec:our-method} we propose our 3D orientation field transform.
 The test electron tomograms data is presented in Section \ref{sec:data}.
 The effectiveness of the proposed method in response to different types of 2D and 3D curves in the test data is detailed in Section \ref{sec:experiment}. Finally, we conclude in Section \ref{sec:oftsummary}.

\section{Mathematical Preliminaries}
\label{sec:preliminaries}
Let $\mathbb{R}^{n}_+ \in \mathbb{R}^{n}$ be a half of the $n$-dimensional Euclidean space generated by an $(n-1)$-dimentional hyperplane crossing the origin. Let $V^n \subset \mathbb{R}^n_+$ be a domain containing all the unit vectors in $\mathbb{R}^n_+$ with the origin as the starting point. Let $\bar{V}^n$ denote the discretized $V^n$, with $|V^n|$ number of unit vectors. For example, $\bar{V}^2$ and $\bar{V}^3$ denote the sets containing unit vectors in the half of the discretized 2D and 3D Euclidean spaces, respectively. 

Let $I$ be an image with domain $\Omega \subset \mathbb{R}^n$. Let $\textbf{x} = (x_1, \cdots, x_n)\in \Omega$ represent individual pixels/voxels of the image $I$. The value of the image $I$ at $\textbf{x}$ is denoted as a function $I(\textbf{x})$. The $l_{2}$-norm of $\textbf{x}$ is denoted by $\lVert\textbf{x}\rVert_{2}=\sqrt{x_{1}^{2} + \cdots + x_n^2}$. 
Although the practical image is discrete, for ease of discussion continuous functions and the integral will be used, which is also common practice in research (e.g. \cite{IEEEhowto:sandberg2007,IEEEhowto:sandberg2009}).

Note that every point e.g. $\textbf{y}\in \mathbb{R}^n$  also corresponds to a unique vector starting from the origin. With abuse of notation, we also call $\textbf{y}$ a vector.  
Given vectors $\textbf{y}, \textbf{z}\in\mathbb{R}^{n}$, the inner product of vectors $\textbf{y}$ and $\textbf{z}$ is represented as $\textbf{y} \cdot \textbf{z}$. The angle between $\textbf{y}$ and $\textbf{z}$ can be calculated by $\arccos(\hat{\textbf{y}} \cdot \hat{\textbf{z}})$, where $\hat{\textbf{y}}$ and $\hat{\textbf{z}}$ are the unit vectors of $\textbf{y}$ and $\textbf{z}$, calculated by $\hat{\textbf{y}}=\textbf{y}/\lVert\textbf{y}\rVert_{2}$ and $\hat{\textbf{z}}=\textbf{z}/\lVert\textbf{z}\rVert_{2}$, respectively. 

\section{Related Work}
\label{sec:relatedwork}
The 2D orientation field transform is a top-down image enhancement method that aims to strengthen curves exclusively. As a long string could be approximated by many small pieces of overlapping line intervals, theoretically, using some sort of line filter of fixed length but of unspecified direction should be able to isolate curves out specifically. 

The first problem, then, is to determine the directions of the line filters at individual pixel $\mathbf{x}$.
The work in \cite{IEEEhowto:sandberg2009} proposed to measure the strength of a line with length $\varepsilon$ centred at $\mathbf{x} \in \Omega \subset \mathbb{R}^2$ along direction $\hat{\mathbf{b}} \in \bar{V}^2$ by a line integral operator ${\cal R}[I]$, i.e.,	
\begin{equation}\label{eq:lineintegral}
 		{\cal R}[I](\mathbf{x},\hat{\mathbf{b}}) = \int _{-{\varepsilon}/{2} }^{{\varepsilon}/{2}}I(\mathbf{x}+s\hat{\mathbf{b}}) ds.
\end{equation}
Obviously, 	
\begin{equation}
    {\cal R}[I](\mathbf{x},\hat{\mathbf{b}}) = 
        {\cal R}[I](\mathbf{x},-\hat{\mathbf{b}}).
\end{equation}
Therefore the direction of $\hat{\mathbf{b}}$ is restricted in $[0, \pi)$, half of the plane, to avoid repetitiveness.  
	
Ways to incorporate the directional information in  ${\cal R}[I]$ have evolved over the course of several papers \cite{IEEEhowto:sandberg2007,IEEEhowto:sandberg2009}.
In \cite{IEEEhowto:sandberg2009}, a primary orientation field at point $\mathbf{x}$, ${\cal F}[{\cal R}](\mathbf{x})$, was generated by taking the maximal line integral ${\cal R}[I]$ of point $\mathbf{x}$ and the direction $\hat{\mathbf{b}}$ achieving this maximal integral, i.e.,
\begin{align} \label{eq:maxintegral}
 		{\cal F}[{\cal R}](\mathbf{x}) 
 		 := & \left\{{\cal F}_1[{\cal R}](\mathbf{x}), \ {\cal F}_2[{\cal R}](\mathbf{x}) \right \} \nonumber \\
 		 = & \Big\{\max_{\hat{\mathbf{b}} \in \bar{V}^2} {\cal R}[I](\mathbf{x},\hat{\mathbf{b}}),  \
 		 \arg\max_{\hat{\mathbf{b}} \in \bar{V}^2} {\cal R}[I](\mathbf{x},\hat{\mathbf{b}})
 		\Big\}.
\end{align}
The alignment integral operator ${\cal G}[{\cal F}]$ at point $\mathbf{x}$ along with direction $\hat{\mathbf{b}}$ reads
\begin{align} 
	\label{eq:alignintegral}
		& {\cal G}[{\cal F}](\mathbf{x},\hat{\mathbf{b}}) \nonumber \\
		= & \int_{-{\varepsilon}/{2}}^{{\varepsilon}/{2}} {\cal F}_1[{\cal R}](\mathbf{x}+s\hat{\mathbf{b}}) \nonumber\\
		& \quad \cos \left(2\arccos({\cal F}_2[{\cal R}](\mathbf{x}+s\hat{\mathbf{b}}) \cdot\hat{\mathbf{b}})\right) ds \nonumber\\
		= & \int_{-{\varepsilon}/{2}}^{{\varepsilon}/{2}}
		{\cal F}_1[{\cal R}](\mathbf{x}+s\hat{\mathbf{b}}) \nonumber \\ 
		& \quad \left(2({\cal F}_2[{\cal R}](\mathbf{x}+s\hat{\mathbf{b}})\cdot\hat{\mathbf{b}})^{2}-1 \right) ds,
\end{align}
which can be used to detect curve-like structures.
The more out of alignment against $\hat{\mathbf{b}}$ a point on the orientation field is, the lower the overall value of the alignment integral would be. In principle, a strong alignment should wind along the length of a curve whilst the opposite should be true for objects which do not have a clear orientation.

The 2D orientation field transform is completed by  taking the maximum value of the alignment with respect to $\hat{\mathbf{b}}$, i.e.,
\begin{equation} \label{eq:maxalign}
{\cal O}[{\cal G}](\mathbf{x})=\mathop{\max}_{\hat{\mathop{\mathbf{b}}}\in \bar{V}^2} {\cal G}[{\cal F}](\mathbf{x},\hat{\mathbf{b}}),
\end{equation}
see \cite{IEEEhowto:sandberg2007} for more details.

\begin{figure}[!tp]
\centering
\begin{tabular}{c}
\includegraphics[width=0.40\textwidth]{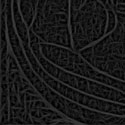} 
\end{tabular}
\caption{Image in Fig. \ref{etexample} enhanced by the 2D orientation field transform with the orientation field defined in \eqref{eq:maxintegral}.}
\label{maxi05compare}
\end{figure}

The orientation field defined in \eqref{eq:maxintegral} is sensitive to non-curve information (e.g. point-like objects) in the given image. For example, in Fig. \ref{maxi05compare}, a slice of an electron tomogram is processed with the previously described orientation field transform \eqref{eq:maxalign} with the orientation field defined in \eqref{eq:maxintegral}. On the one hand, the curves in the given image are 
successfully amplified. On the other hand, the structures around small dots in the given image are not suppressed but enhanced as curves. To overcome this issue raised by the orientation field defined in \eqref{eq:maxintegral}, a new orientation field, an average orientation, was defined in \cite{IEEEhowto:sandberg2009}. The average orientation was then used in equation \eqref{eq:maxalign} to form the 2D orientation field transform. It is worth mentioning that the 2D orientation field transform proposed in \cite{IEEEhowto:sandberg2009} is extremely computationally expensive. We refer the readers to \cite{IEEEhowto:sandberg2009} for more details. 
Note, importantly, that the above mentioned 2D orientation field transforms are focusing on 2D space and are all less generalisable to higher dimensions.

\section{Proposed 3D orientation field transform}
\label{sec:our-method}
As evidenced in \cite{IEEEhowto:sandberg2009}, the average operation noticeably improved discrimination of curves from other structures compared to \cite{IEEEhowto:sandberg2007}. Nevertheless, there is no 3D analogue where vectors from exactly half of a Euclidean space could be transformed bijectively to cover exactly the entirety of Euclidean space. Still, the idea of detecting the directionality of a neighbourhood would be inspirational. We proposed the 3D orientation field transform below. 

Firstly, the line integral operator ${\cal R}[I]$, the orientation field 
\begin{equation} 
{\cal F}[{\cal R}] := \left\{{\cal F}_1[{\cal R}](\mathbf{x}), \ {\cal F}_2[{\cal R}](\mathbf{x}) \right \},
\end{equation}
and the orientation field transform ${\cal O}[{\cal G}]$ respectively defined in \eqref{eq:lineintegral}, \eqref{eq:maxintegral} and \eqref{eq:maxalign} are extended from 2D to 3D, i.e., for $\forall\mathbf{x} \in \Omega \subset \mathbb{R}^3$ and $\forall\hat{\mathbf{b}} \in \bar{V}^3$, then
\begin{align}
 		{\cal R}[I](\mathbf{x},\hat{\mathbf{b}}) & = \int _{-{\varepsilon}/{2}}^{{\varepsilon}/{2}}I(\mathbf{x}+s\hat{\mathbf{b}}) ds, \label{eq:lineintegral-3d}\\
 		{\cal F}_1[{\cal R}](\mathbf{x}) & = \max_{\hat{\mathbf{b}} \in \bar{V}^3} {\cal R}[I](\mathbf{x},\hat{\mathbf{b}}), \label{eq:maxintegral-3d-1} \\ 
        {\cal F}_2[{\cal R}](\mathbf{x}) & = \arg\max_{\hat{\mathbf{b}} \in \bar{V}^3} {\cal R}[I](\mathbf{x},\hat{\mathbf{b}}), 
        \label{eq:maxintegral-3d-2} \\
        {\cal O}[{\cal G}](\mathbf{x}) & =\mathop{\max}_{\hat{\mathop{\mathbf{b}}}\in \bar{V}^3} {\cal G}[{\cal F}](\mathbf{x},\hat{\mathbf{b}}), \label{eq:maxalign-3d}
\end{align}
where ${\cal G}[{\cal F}](\mathbf{x},\hat{\mathbf{b}})$ is the alignment integral operator given in \eqref{eq:alignintegral}, i.e.,
\begin{align} 
	\label{eq:alignintegral-3d}
		& {\cal G}[{\cal F}](\mathbf{x},\hat{\mathbf{b}}) \nonumber \\
		= & \int_{-{\varepsilon}/{2}}^{{\varepsilon}/{2}}
		{\cal F}_1[{\cal R}](\mathbf{x}+s\hat{\mathbf{b}}) \Big(2({\cal F}_2[{\cal R}](\mathbf{x}+s\hat{\mathbf{b}})\cdot\hat{\mathbf{b}})^{2} 
		  -1 \Big) ds.
\end{align}  
Recall that the line integral operator ${\cal R}[I](\mathbf{x},\hat{\mathbf{b}})$ of the image $I$ measures the strength of each line with length $\varepsilon$ centred at $\mathbf{x} \in \Omega$ along direction $\hat{\mathbf{b}} \in {\bar V}^3$.

The orientation transform ${\cal O}[{\cal G}]$ equipped with the maximum of the line integrals ${\cal F}_1[{\cal R}]$ for 2D can enhance curves but suffer from for example point-like objects \cite{IEEEhowto:sandberg2009}. Therefore, it is not enough to  use ${\cal O}[{\cal G}]$ defined in \eqref{eq:maxalign-3d} to directly enhance curves in 3D which is much more complicated than the case in 2D. The main issue of only using the maximum to identify the curve direction is that it disregards the number of large line integrals running along different directions at a point; in other words, the maximum criterion in this scenario will mistakenly judge the points e.g. inside point-like objects to be on a curve.

It is clear that, in a neighbourhood of a point that shows a clear orientation, the integral along this direction will be of a fairly higher value than the others. On the contrary, in a neighbourhood of a point that is centred on a point-like object or covered with a homogeneous signal, integrals along one direction should have little difference to the others. Hence, measuring the magnitude and variability of integrals at a point should be indicative of whether its neighbourhood encloses a curve or not. 
Since the mean and absolute deviation (acting as low-pass and high-pass filters, respectively) are powerful to estimate this kind of variability, to overcome the challenge above, the mean and absolute deviation of the set of the line integral values $\{{\cal R}[I](\mathbf{x},\hat{\mathbf{b}})\}_{\hat{\mathbf{b}} \in \bar{V}^3}$ and the set of the alignment integral values $\{{\cal G}[{\cal F}](\mathbf{x},\hat{\mathbf{b}})\}_{\hat{\mathbf{b}} \in \bar{V}^3}$ will be introduced to design our 3D orientation field transform. 

\begin{remark}
Similar to the maximum of the line and alignment integrals,
the means of the line and alignment integrals may also be non-selective to point-like objects. The difference between the mean and the maximum is that the mean averages out the signal along different directions, effectively acting as a low-pass filter.
\end{remark}

The mean and the absolute deviation of the set of the line integral values $\{{\cal R}[I](\mathbf{x},\hat{\mathbf{b}})\}_{\hat{\mathbf{b}} \in \bar{V}^3}$ are defined as
\begin{equation}
\label{eq:sumr}
{\cal M}[{\cal R}](\mathbf{x}) = \frac{1}{|\bar{V}^3|} \sum_{\hat{\mathbf{b}} \in \bar{V}^3}{\cal R}[I](\mathbf{x},\hat{\mathbf{b}})
\end{equation}
and
\begin{equation}
\label{eq:l1r}
\sigma[{\cal R}](\mathbf{x}) = \frac{1}{|\bar{V}^3|}
\sum_{\hat{\mathbf{b}} \in \bar{V}^3}\vert {\cal M}[{\cal R}](\mathbf{x}) - {\cal R}[I](\mathbf{x},\hat{\mathbf{b}})\vert
\end{equation}
respectively. 
Analogously, the mean and the absolute deviation of the set of the alignment integral values $\{{\cal G}[{\cal F}](\mathbf{x},\hat{\mathbf{b}})\}_{\hat{\mathbf{b}} \in \bar{V}^3}$ are defined as
\begin{equation}
\label{eq:sumi}
{\cal M}[{\cal G}](\mathbf{x}) = \frac{1}{|\bar{V}^3|} \sum_{\hat{\mathbf{b}} \in \bar{V}^3}{\cal G}[{\cal F}](\mathbf{x},\hat{\mathbf{b}}),
\end{equation}
and
\begin{equation}
\label{eq:l1i}
\sigma[{\cal G}](\mathbf{x}) = \frac{1}{|\bar{V}^3|} \sum_{\hat{\mathbf{b}} \in \bar{V}^3}\vert {\cal M}[{\cal G}](\mathbf{x}) - {\cal G}[{\cal F}](\mathbf{x},\hat{\mathbf{b}})\vert,
 \end{equation}
respectively.

\begin{remark}
Before computing the mean and absolute deviation defined in \eqref{eq:sumr}--\eqref{eq:l1i}, the values in each set of $\{{\cal R}[I](\mathbf{x},\hat{\mathbf{b}})\}_{\hat{\mathbf{b}} \in \bar{V}^3}$ and $\{{\cal G}[{\cal F}](\mathbf{x},\hat{\mathbf{b}})\}_{\hat{\mathbf{b}} \in \bar{V}^3}$ can also be polished using some smoothing operators like Gaussian. 
\end{remark}

\begin{figure*}[!tp]
	\centering
	\begin{tabular}{ccc}
		\includegraphics[width=0.28\textwidth]{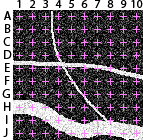} &
		\includegraphics[width=0.28\textwidth]{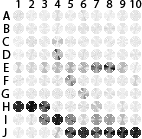} & 
		\includegraphics[width=0.28\textwidth]{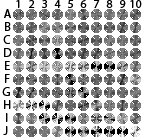} \\
		(a) & (b) & (c) \\
        & 
        \includegraphics[width=0.28\textwidth]{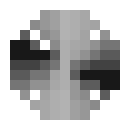} & \includegraphics[width=0.28\textwidth]{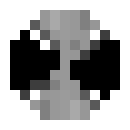}
		\\
		& (d) & (e) 	 
    \end{tabular}
\caption{Demonstration in 2D of the information extracted out of the line integral ${\cal R}[I]$ and alignment integral ${\cal G}[{\cal F}]$. (a): a binary test image peppered with a light layer of Gaussian noise, where the cross-hairs mark the selected pixels used in (b)--(c); (b): distributions of line integral values at each direction for selected pixels in (a);  (c): distributions of alignment integral values at each direction for selected pixels in (a); and (d)--(e): close up of the disc E8 in (b) and (c). In particular, intensity values in (b)--(e) are normalised to the range of [0, 1], where white and dark colours depict the lowest and highest integral values, respectively. }
\label{oftdemonstration}
\end{figure*}

Fig. \ref{oftdemonstration} demonstrates the characteristics of the line integrals (i.e., eqn  \eqref{eq:lineintegral-3d}) and alignment integrals (i.e., eqn \eqref{eq:alignintegral-3d}), which the maximum, mean and absolute deviation measures (i.e., \eqref{eq:maxintegral-3d-1}, \eqref{eq:maxalign-3d}, \eqref{eq:sumr}--\eqref{eq:l1i}) are built on. 
Fig. \ref{oftdemonstration} (a) is the test image with selected pixels marked as cross-hairs. Each disc in Fig. \ref{oftdemonstration} (b) and (c) corresponds to one pixel and represents the distribution of the integrals values in every direction, see \ref{oftdemonstration} (d) and (e) for two close-up discs. The darker the line, the higher the integral value. 
Note that the demonstration of Fig. \ref{oftdemonstration} is done in 2D for the purpose of better visualisation. The demonstration in higher dimensions like 3D is in the same fashion.

The discs in Fig. \ref{oftdemonstration} (b) and (c) 
can disclose which pixels have a high maximum, mean and absolute deviation of the line and alignment integrals, i.e. the pixels on or off the curve structures. For example, in Fig. \ref{oftdemonstration} (b), discs D4, E8 and H1, which show the line integrals of the corresponding three pixels on the curve in Fig. \ref{oftdemonstration} (a), indeed possess high maximum, mean or absolute deviation. A similar conclusion can also be seen from Fig. \ref{oftdemonstration} (c). On the whole, pixels on a curve have an overall higher maximum, mean and/or absolute deviation than those that are off a curve.

For simplicity, let 
\begin{align*}
    {\cal W}_1(\mathbf{x}) & = {\cal F}_1[{\cal R}](\mathbf{x}), 
    & 
    {\cal W}_2(\mathbf{x}) & = {\cal O}[{\cal G}](\mathbf{x}), \nonumber \\
    {\cal W}_3(\mathbf{x}) & = {\cal M}[{\cal R}](\mathbf{x}), 
    &
    {\cal W}_4(\mathbf{x}) & = {\cal M}[{\cal G}](\mathbf{x}), \nonumber \\
    {\cal W}_5(\mathbf{x}) & = \sigma[{\cal R}](\mathbf{x}), 
    &
    {\cal W}_6(\mathbf{x}) & = \sigma[{\cal G}](\mathbf{x}). \nonumber
\end{align*} 
Finally, our proposed 3D orientation field transform ${\cal O}_{\rm 3D}[I](\mathbf{x})$ are constructed by leveraging all the  measures -- the maximum, mean and absolute deviation of the line integral and alignment integral -- to detect curves in 3D images, i.e., 
\begin{equation}
\label{eq:firstoft}
    {\cal O}_{\rm 3D}[I](\mathbf{x}) = f(\{{\cal W}_i(\mathbf{x})\}_{i=1}^6),
\end{equation}
where $f$ is a function with the six measures as inputs. In this paper, the forms of 
\begin{align}
f(\{{\cal W}_i(\mathbf{x})\}_{i=1}^6) & = \Pi_{i=1}^6 {\cal W}_i(\mathbf{x}),  \label{3d-oft-f-all}  \\
f(\{{\cal W}_i(\mathbf{x})\}_{i=1}^6) & = \Pi_{i=1, i\neq4}^6 {\cal W}_i(\mathbf{x}), \label{3d-oft-f-no-4} \\
f(\{{\cal W}_i(\mathbf{x})\}_{i=1}^6) & =  {\cal W}_1(\mathbf{x}) {\cal W}_3(\mathbf{x}),  \label{3d-oft-f-1-3}
\end{align}
are considered. We leave other choices of $f$ for future investigation. 

The 3D orientation field transform is summarised in Algorithm \ref{alg:3d-oft}. It is worth remarking that the above proposed 3D orientation field transform can naturally be extended to any dimensions. 

\begin{algorithm} 
\caption{3D orientation field transform}
\label{alg:3d-oft}
 \textbf{Input:} 3D image $I(\mathbf{x})$, $\mathbf{x}\in \Omega$ \\
 \textbf{Output:} Curve-enhanced image ${\cal O}_{\rm 3D}[I]$ \vspace{0.05in} \\ 
 		Compute the orientation field ${\cal F}_1[{\cal R}](\mathbf{x})$ using eqn \eqref{eq:maxintegral-3d-1};
		\\
		Compute the orientation field transform 
		${\cal O}[{\cal G}](\mathbf{x})$ using eqn \eqref{eq:maxalign-3d};
		\\
        Compute the mean of the set of the line integral values ${\cal M}[{\cal R}](\mathbf{x})$ using eqn \eqref{eq:sumr};
        \\
        Compute the absolute deviation of the set of the line integral values $\sigma[{\cal R}](\mathbf{x})$ using eqn \eqref{eq:l1r};
		\\
		Compute the mean of the set of the line integral values ${\cal M}[{\cal G}](\mathbf{x})$ using eqn \eqref{eq:sumi};
        \\
        Compute the absolute deviation of the set of the alignment integral values $\sigma[{\cal G}](\mathbf{x})$ using eqn \eqref{eq:l1i};
		\\
		Compute the 3D orientation field transform ${\cal O}_{\rm 3D}[I](\mathbf{x})$ using eqn \eqref{eq:firstoft}.
\end{algorithm}

\section{Test data}
\label{sec:data}
Common electron microscopy protocols use heavy metal compounds, namely osmium tetroxide, uranyl acetate and lead citrate as staining agents that adsorb on macromolecular complexes in the biological sample. As a typical cell is made mostly of light atoms, these heavy metal conjugates are responsible for deflecting the electrons to generate image contrast.
	
The protocol used to create the images here used freeze-substitution instead of chemical fixation for immobilising subcellular structures in a soft solid in preparation for embedding it in hard resin for imaging. The advantage of using freeze-substitution is that it prevents the distortion of intra-cellular architecture during infiltration of chemical cross-linkers and dehydration for resin embedding, that the standard chemical fixation protocols involve. However, samples processed with chemical fixation have a higher signal-to-noise ratio as the chemical fixatives collapse macromolecule to which heavy metal stain concentrates. Furthermore, the cytosol and organelle lumen are washed away during dehydration, leaving empty backgrounds. As a result, subcellular structures in the electron micrographs used here are not distinguished so much as those in conventional electron micrographs.
	
All samples used for the test images (Fig. \ref{origimages}) in this paper were imaged using electron tomography, which is a computational tomography version of transmission electron microscopy. Scanning transmission electron microscopy was used instead of transmission electron microscopy as a sub-process. The former uses a raster scanning method while the latter does not. Therefore the former would improve the image resolution. For the computational tomography, two series of images were taken for each sample by sequentially tilting the sample along two orthogonal directions with an angular difference of 1.5$^{\circ}$ ~each up to a maximum of $\pm$60$^{\circ}$. Then the simultaneous iterative reconstruction technique (SIRT) developed by \cite{gilbert1972iterative} and adapted in IMOD was used to reconstruct the 3D tomograms using those images.
	
The samples were tilted only up to $\pm$60$^{\circ}$ ~as otherwise, the paths of the electrons would become too long for them to pass through, since electrons are very reactive to any matter. Hence, it is a compromise between the sample thickness and the maximum imaging angle. However, that would create a missing-wedge problem, where reconstructed images were blurred along the $z$-axis, complicating the curve enhancement and segmentation of any 3D structures. More details are presented in the Appendix.

\section{Experimental results}
	\label{sec:experiment}
\begin{figure*}[!tp]
	\centering
	\begin{tabular}{cccc}
	    & Maximum & Mean  & Absolute deviation \vspace{0.08in}
	    \\
	    \rotatebox{90}{\hspace{0.15in} Line integral ${\cal R}[I]$} &
	    \includegraphics[width=0.27\textwidth]{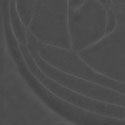} &
		\includegraphics[width=0.27\textwidth]{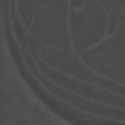} &
		\includegraphics[width=0.27\textwidth]{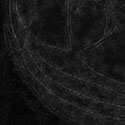} \\
		& (a) & (b) & (c) \\
	    \rotatebox{90}{\hspace{0.0in} Alignment integral ${\cal G}[{\cal F}]$} &
	    \includegraphics[width=0.27\textwidth]{./media/maxi05.jpeg} & 
		\includegraphics[width=0.27\textwidth]{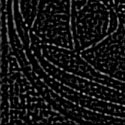} &
		\includegraphics[width=0.27\textwidth]{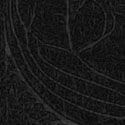} \\
		& (d) & (e) & (f) 	 
    \end{tabular}
\caption{Maximum, mean and absolute deviation of the line integral ${\cal R}[I]$ and alignment integral ${\cal G}[{\cal F}]$ on test image in Fig. \ref{origimages} (a). Columns from left to right respectively give the maximum, mean and absolute deviation of the line integral (first row) and alignment integral (second row). }
\label{2dmodules-one}
\end{figure*}

\begin{figure*}[!tp]
	\centering
	\begin{tabular}{cccc}
	    & Maximum & Mean  & Absolute deviation \vspace{0.08in}
	    \\
	    \rotatebox{90}{\hspace{0.15in} Line integral ${\cal R}[I]$} &
	    \includegraphics[width=0.27\textwidth]{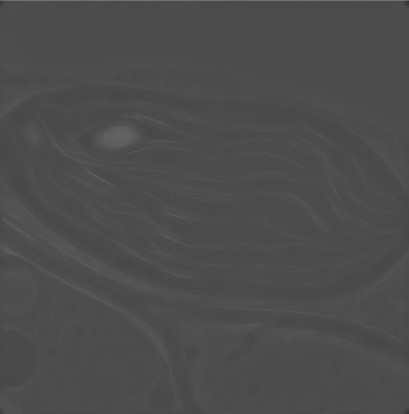} &
		\includegraphics[width=0.275\textwidth]{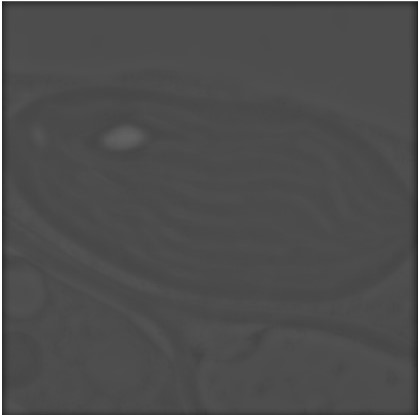} &
		\includegraphics[width=0.27\textwidth]{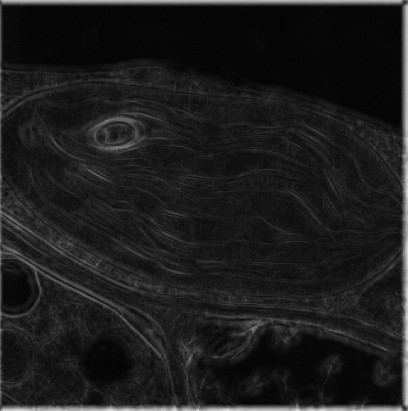} \\
		& (a) & (b) & (c) \\
	    \rotatebox{90}{\hspace{0.0in} Alignment integral ${\cal G}[{\cal F}]$} &
	    \includegraphics[width=0.27\textwidth]{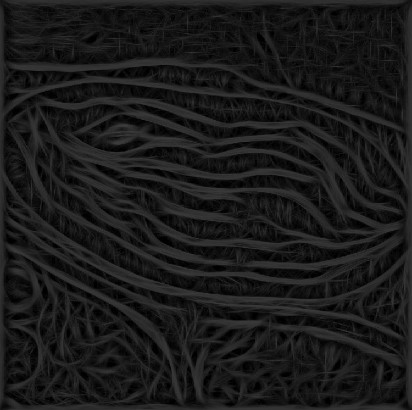} & 
		\includegraphics[width=0.27\textwidth]{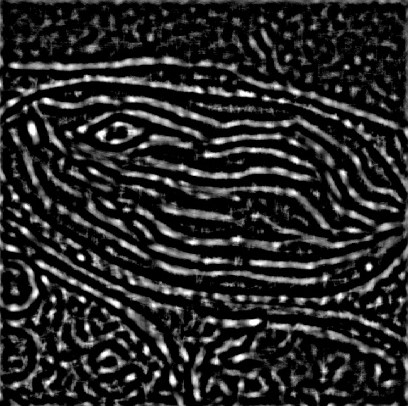} &
		\includegraphics[width=0.27\textwidth]{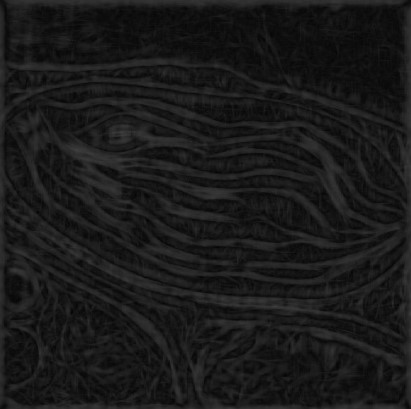} \\
		& (d) & (e) & (f) 	 
    \end{tabular}
\caption{Maximum, mean and absolute deviation of the line integral ${\cal R}[I]$ and alignment integral ${\cal G}[{\cal F}]$ on test image in Fig. \ref{origimages} (b). Columns from left to right respectively give the maximum, mean and absolute deviation of the line integral (first row) and alignment integral (second row). }
\label{2dmodules-two}
\end{figure*}

Three real-world images shown in Fig. \ref{origimages} were tested with the proposed 3D orientation field transform. The first one (Fig. \ref{origimages} (a)) is an image containing sparse 2D curves; the second one (Fig. \ref{origimages} (b)) is an image with densely packed and heterogeneously stacked 2D curves with varying thickness; and the last one (Fig. \ref{origimages} (c)--(d)) is a 3D image of interconnecting 3D curves, which is extremely challenging. The one in Fig. \ref{origimages} (e) is a synthetic mesh of a 3D curve among point-like objects made for a demonstration of the proposed 3D orientation field transform. 

There is only one parameter to be set in the proposed 3D orientation field transform, which is the length $\varepsilon$ of the paths for the integral. Motivated by the estimation in \cite{IEEEhowto:sandberg2007,IEEEhowto:sandberg2009}, the length is set to be 1.5 times of the thickness of the curve to be enhanced in order for all curves to be identified properly as a curve rather than a surface.

The proposed 3D orientation field transform is experimented first with 2D images shown in Fig. \ref{origimages} (a)--(b) since i) 2D image can be regarded as a special case of a 3D image; and ii) it is easier to demonstrate curve enhancement on 2D images  than 3D ones. After that, the proposed transform will be evaluated on the synthetic 3D curve in  Fig. \ref{origimages} (e) and the 3D image with 3D curves shown in Fig. \ref{origimages} (c)--(d).

\subsection{Performance in 2D}
	\label{sec:experiment-2d}
\begin{figure}[!tp]
\centering
\begin{tabular}{cc}
\includegraphics[width=0.22\textwidth]{./media/orig05.png} \\
(a) 
\end{tabular}
\begin{tabular}{cc}
\includegraphics[width=0.22\textwidth]{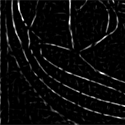} & 
\includegraphics[width=0.22\textwidth]{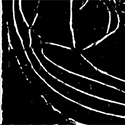}\\
 (b)  & (c) \\
\includegraphics[width=0.22\textwidth]{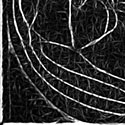} & 
\includegraphics[width=0.22\textwidth]{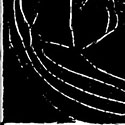}\\
 (d) & (e) 
\end{tabular}
\caption{Performance of the proposed orientation filed transform on test image in Fig. \ref{origimages} (a).
(a): given image; (b) and (d): results of the proposed orientation filed transform using $f$ defined in eqn \eqref{3d-oft-f-all} and \eqref{3d-oft-f-no-4}, respectively; (c) and (e): the segmentation results obtained by hard thresholding of (b) and (d), respectively. }
\label{v1}
\end{figure}

\begin{figure}[!tp]
\centering
\begin{tabular}{c}
\includegraphics[width=0.22\textwidth]{./media/orig12.jpeg} \\
(a)
\end{tabular}
\begin{tabular}{cc}
 \includegraphics[width=0.222\textwidth]{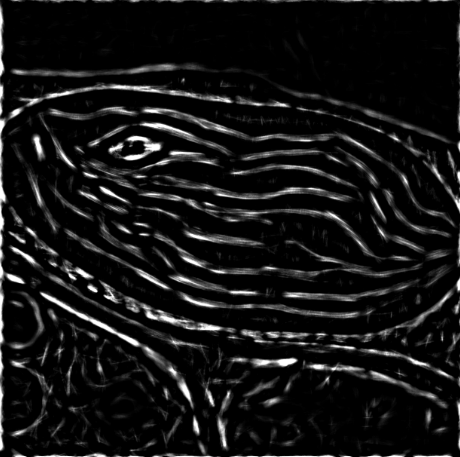} & 
\includegraphics[width=0.22\textwidth]{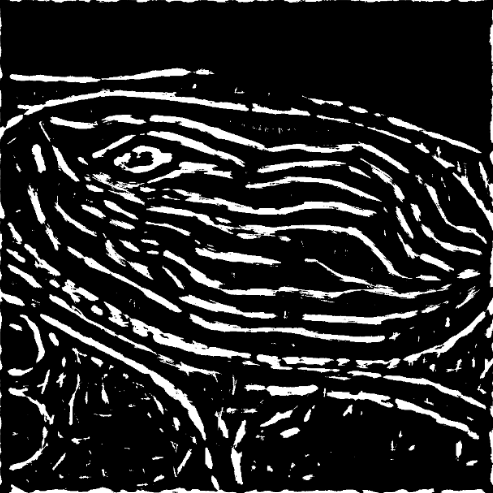} \\
(b)  & (c) \\
\includegraphics[width=0.22\textwidth]{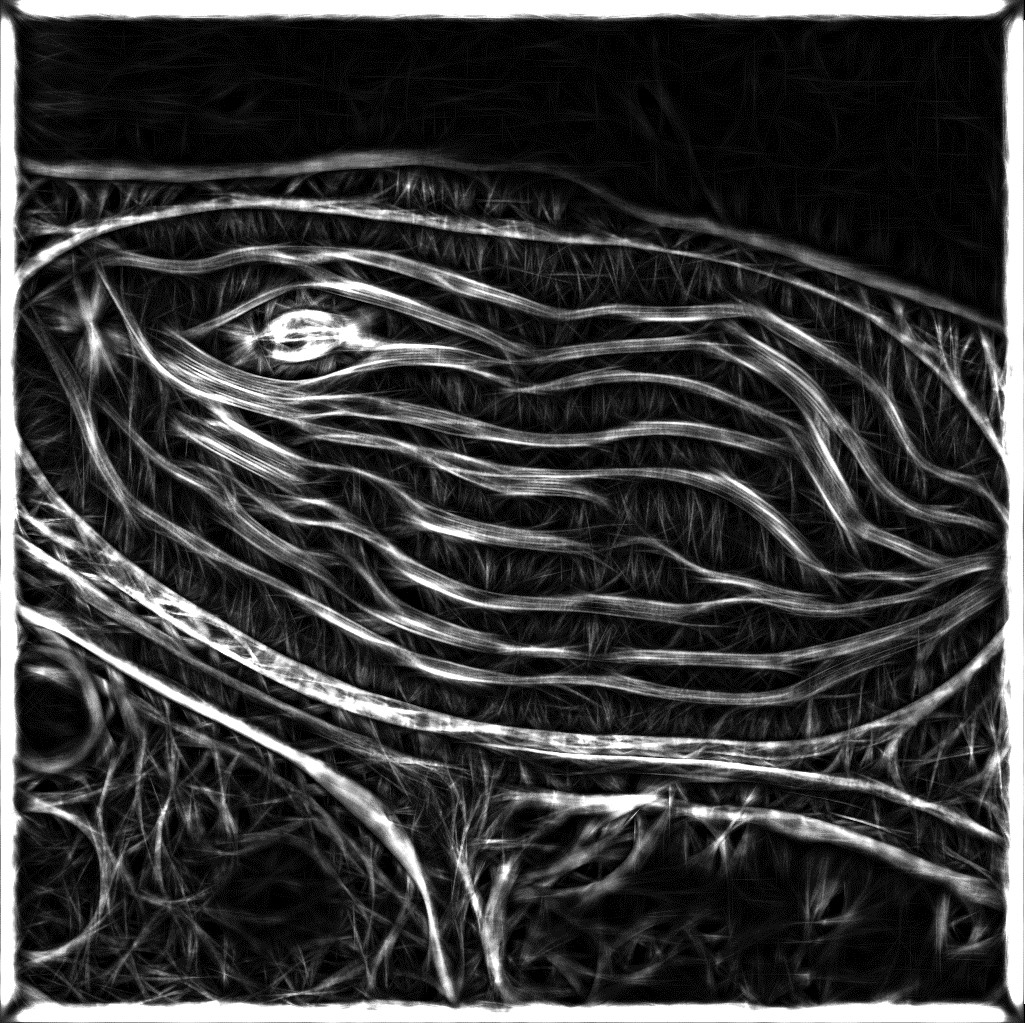} & 
\includegraphics[width=0.22\textwidth]{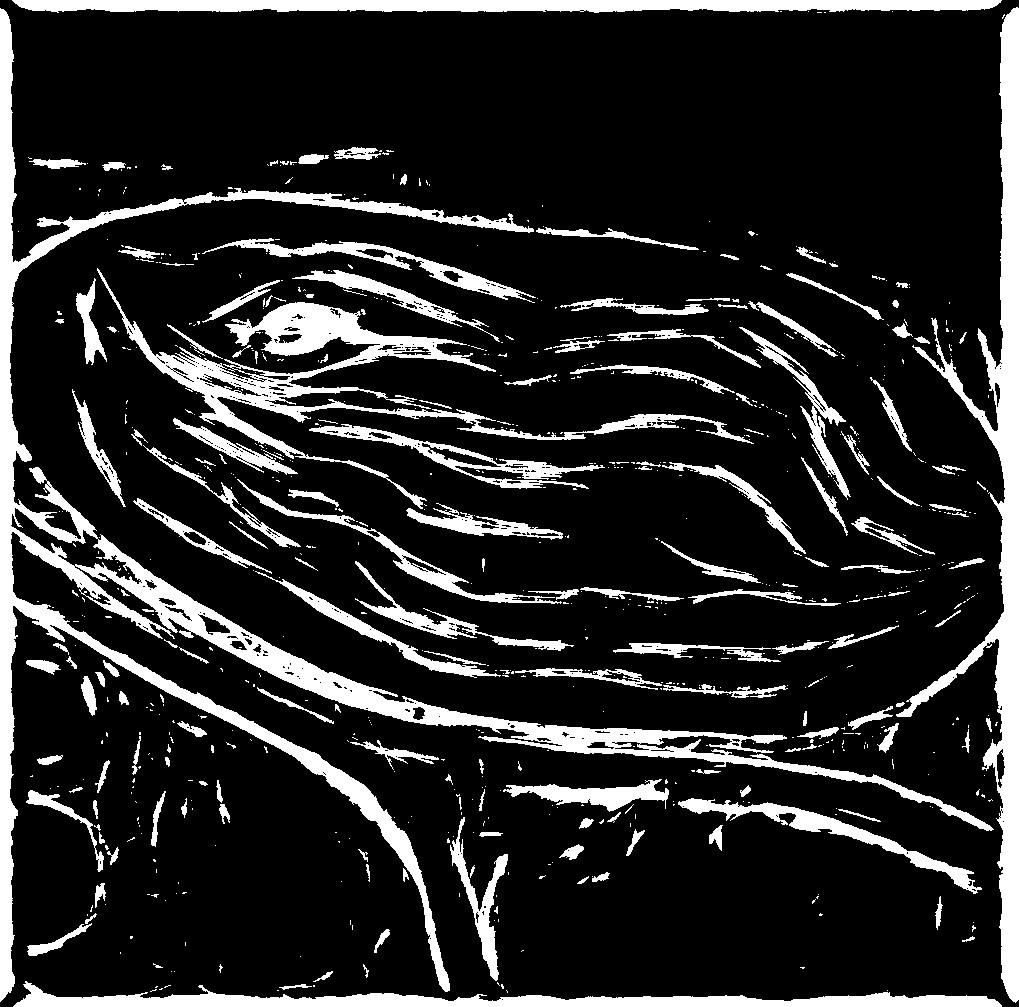} \\
(d) & (e) \\
\end{tabular}
\caption{Performance of the proposed orientation filed transform on test image in Fig. \ref{origimages} (b).
(a): given image; (b) and (d): results of the proposed orientation filed transform using $f$ defined in eqn \eqref{3d-oft-f-all} and \eqref{3d-oft-f-no-4}, respectively; (c) and (e): the segmentation results obtained by hard thresholding of (b) and (d), respectively. }
\label{v2}
\end{figure}

Firstly, the performance of the six measures of the maximum, mean and absolute deviation of the line integral ${\cal R}[I]$ and alignment integral ${\cal G}[{\cal F}]$ are presented in Fig. \ref{2dmodules-one} and Fig. \ref{2dmodules-two}. The results show that the measures of the maximum and mean of the line integral (i.e., eqn \eqref{eq:maxintegral-3d-1} and eqn \eqref{eq:sumr}) perform similarly, acting as generic low-pass filters with no distinctly selective curve enhancement, see (a)--(b) in Fig. \ref{2dmodules-one} and Fig. \ref{2dmodules-two}. Nevertheless, the measures of the maximum and mean of the alignment integral (i.e., eqn \eqref{eq:maxalign-3d} and eqn \eqref{eq:sumi}) can both enhance curves but perform slightly differently, i.e., the mean of the alignment integral achieves results with higher contrast but much noisier than that of the maximum, see (d)--(e) in Fig. \ref{2dmodules-one} and Fig. \ref{2dmodules-two}. In particular, the retained curves using the 
mean of the alignment integral seemingly undulated in intensity in the manner of self-interfering waves, see (e) in Fig. \ref{2dmodules-one} and Fig. \ref{2dmodules-two}. 
The measures of the absolute deviation of the line integral and alignment integral can both enhance the curves and suppress non-curve structures like the light-coloured blobs successfully, even though the results of the absolute deviation of the line integral is slightly blurry compared to the results of the absolute deviation of the alignment integral, see (c) and (f) in Fig. \ref{2dmodules-one} and Fig. \ref{2dmodules-two}. 

The efficacy of combining the above mentioned six transform components through the function in eqn \eqref{eq:firstoft}, i.e., Alg. \ref{alg:3d-oft} with function $f$ defined in eqn \eqref{3d-oft-f-all}, is shown in (b) of Fig. \ref{v1} and Fig. \ref{v2}. We see that the curves are enhanced successfully with the background information suppressed excellently. The only debate is some of the curves are unnecessarily fragmented, which might be caused by the mean of alignment integrals ${\cal M}[{\cal G}](\mathbf{x})$ (see (e) in Fig. \ref{2dmodules-one} and Fig. \ref{2dmodules-two}) as discussed before. In (d) of Fig. \ref{v1} and Fig. \ref{v2}, the results of the proposed transform without using the mean of alignment integrals (i.e., $f$ defined in \eqref{3d-oft-f-no-4}) show that the unnecessarily fragmented curves present in (b) of Fig. \ref{v1} and Fig. \ref{v2} are indeed repaired. In (c) and (e) of Fig. \ref{v1} and Fig. \ref{v2}, the segmentation results obtained by hard thresholding of (b) and (d) of Fig. \ref{v1} and Fig. \ref{v2} validate that the proposed transform can also be served as a tool for example for segmentation, where the thresholding value is selected as a compromise between removing the noise and keeping the curves. 

\begin{figure*}[!tp]
	\centering
	\begin{tabular}{cccc}
	    & Maximum & Mean  & Absolute deviation \vspace{0.08in}
	    \\
	    \rotatebox{90}{\hspace{0.15in} Line integral ${\cal R}[I]$} &
	    \includegraphics[width=0.25\textwidth]{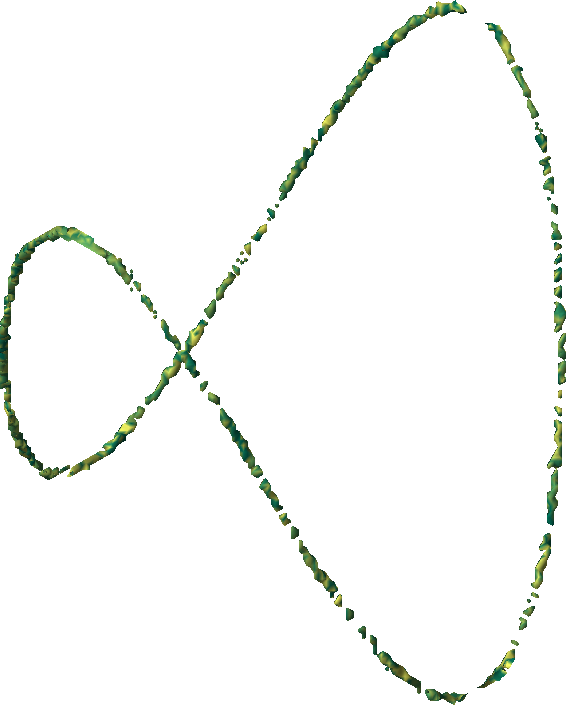} &
		\includegraphics[width=0.25\textwidth]{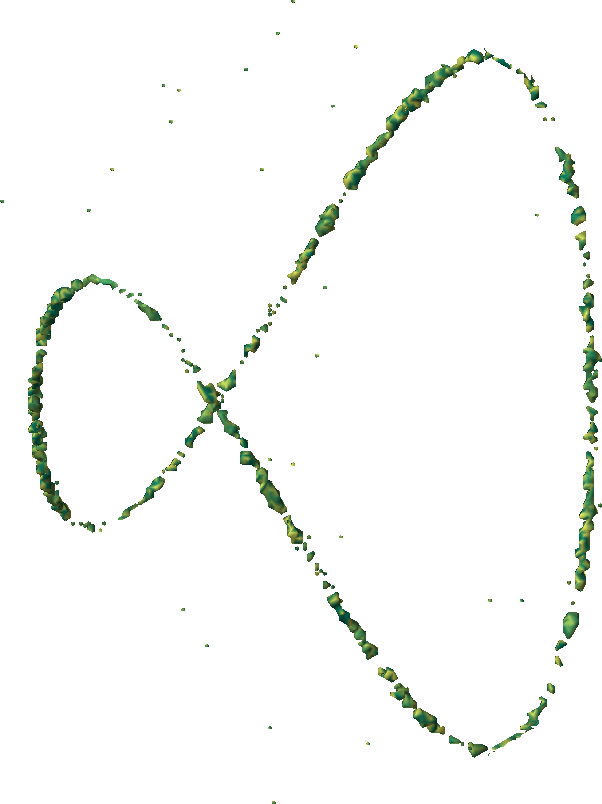} &
		\includegraphics[width=0.28\textwidth]{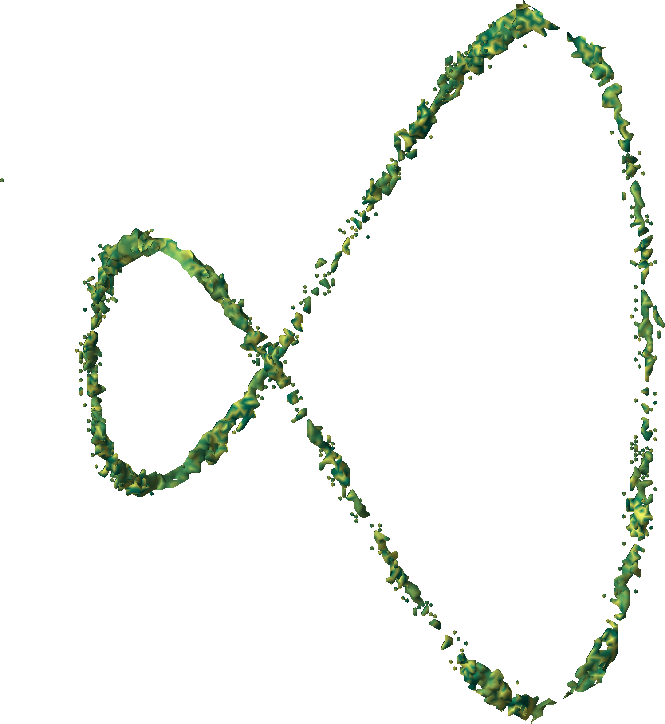} \\
		& (a) & (b) & (c) \\
	    \rotatebox{90}{\hspace{0.0in} Alignment integral ${\cal G}[{\cal F}]$} &
	    \includegraphics[width=0.25\textwidth]{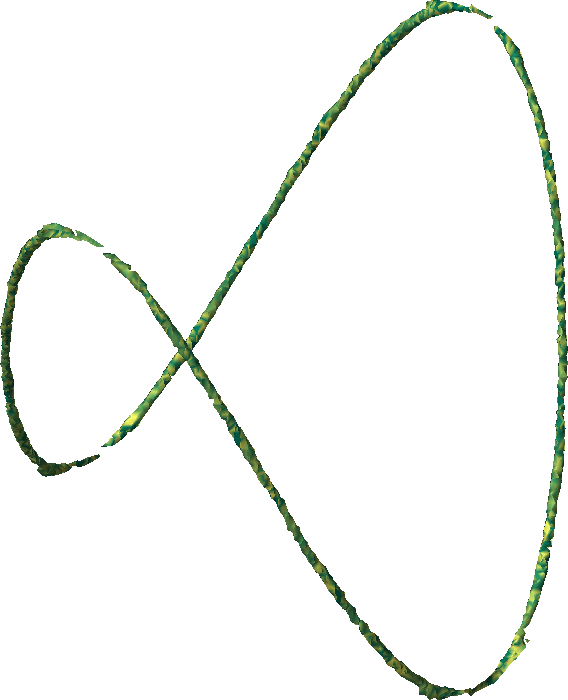} & 
		\includegraphics[width=0.25\textwidth]{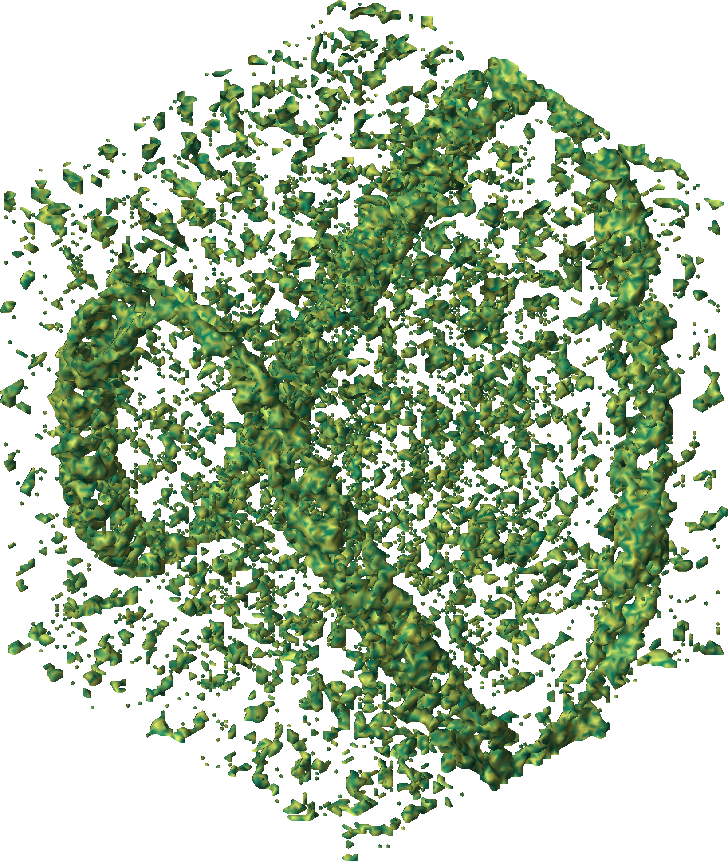} &
		\includegraphics[width=0.25\textwidth]{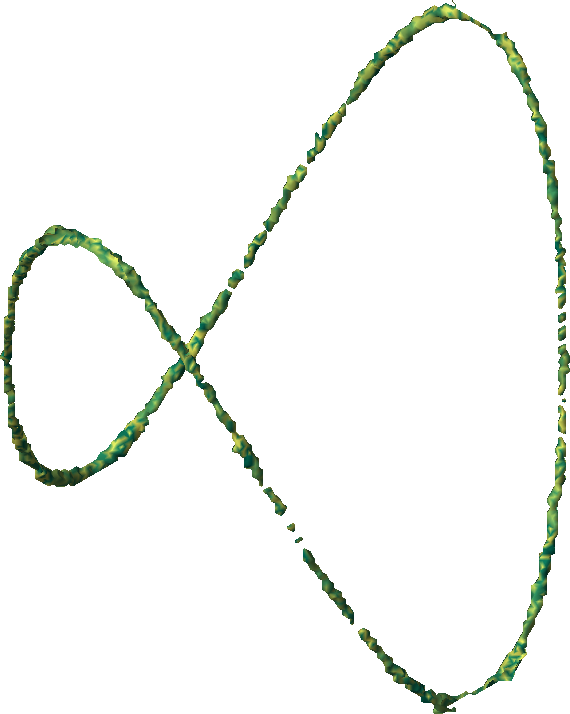} \\
		& (d) & (e) & (f) 	 
    \end{tabular}
\caption{Maximum, mean and absolute deviation of the line integral ${\cal R}[I]$ and alignment integral ${\cal G}[{\cal F}]$ on the synthetic 3D image in Fig. \ref{origimages} (e). Columns from left to right respectively give the maximum, mean and absolute deviation of the line integral (first row) and alignment integral (second row). }
\label{3dsimplecurve}
\end{figure*}

\begin{figure*}[!tp]	
\centering
    \begin{tabular}{ccc}
	    \includegraphics[width=0.28\textwidth]{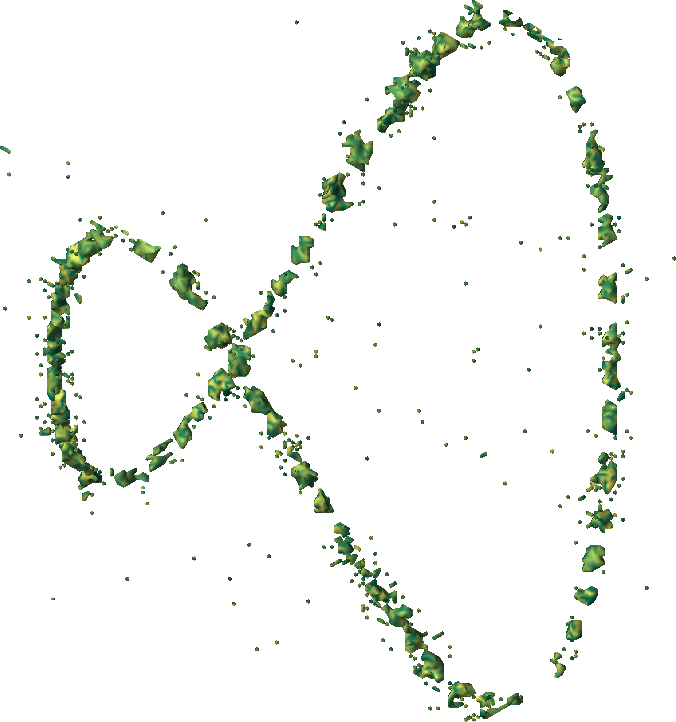} &
		\includegraphics[width=0.25\textwidth]{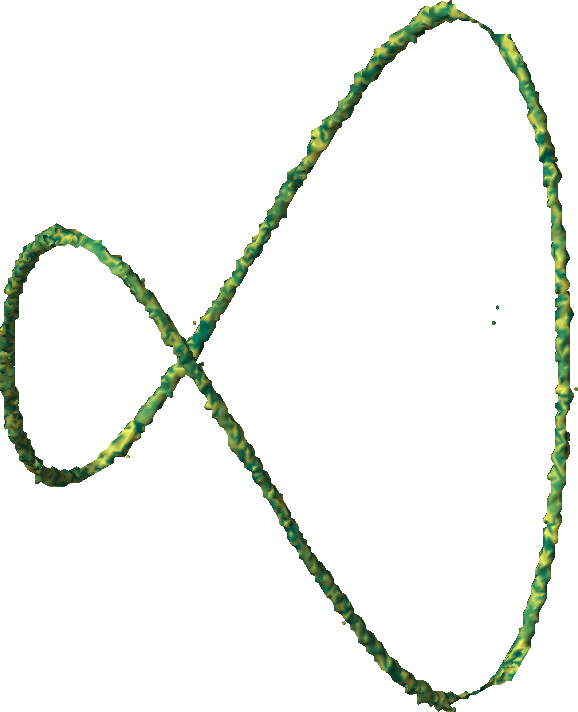} &
		\includegraphics[width=0.26\textwidth]{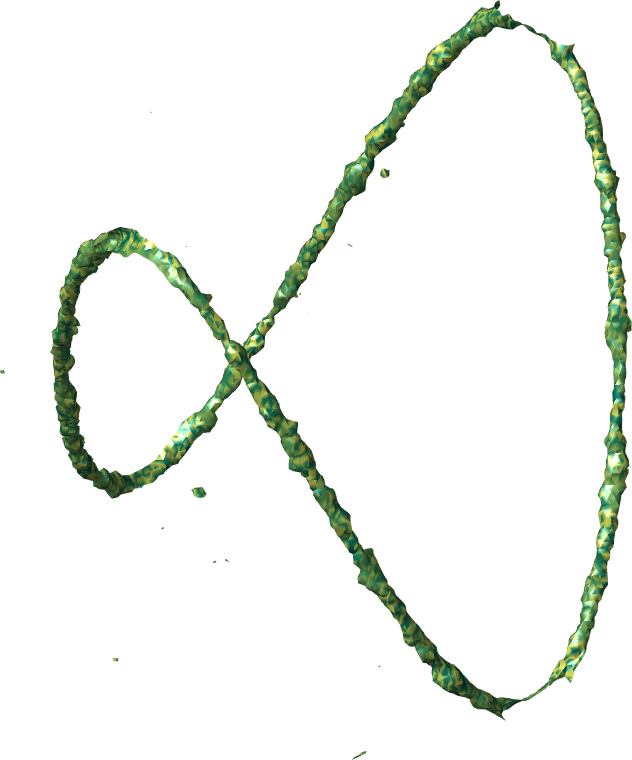} \\
		(a) & (b) & (c) 	 
    \end{tabular}
\caption{Performance of the proposed orientation field transform on the synthetic 3D image in Fig. \ref{origimages} (e). (a)--(c): segmentation results obtained by hard thresholding of the enhanced curves using function $f$ in eqn \eqref{3d-oft-f-all}, \eqref{3d-oft-f-no-4} and \eqref{3d-oft-f-1-3}, respectively. }
\label{3dcomponents}
\end{figure*}

\subsection{Performance in 3D}
	\label{sec:experiment-3d}
\subsubsection{Synthetic 3D image}
The proposed method is first tested on the synthetic 3D image in Fig. \ref{origimages} (e). In order to display 3D images here, the meshes are computed with an adaptation of the Marching Cubes algorithm \cite{IEEEhowto:hammer} for MATLAB and are displayed with a built-in MATLAB GUI. The extremely dense point-like objects surrounding the 3D curve make the curve enhancement and detection very challenging. The performance of the measures of the maximum, mean and absolute deviation of the line integral ${\cal R}[I]$ and alignment integral ${\cal G}[{\cal F}]$ are presented in Fig. \ref{3dsimplecurve}. It shows clearly that all the measures are able to enhance the curves except for the mean of the alignment integral. The segmentation results obtained by hard thresholding of the enhanced curves using the proposed transform with different function $f$ are shown in Fig. \ref{3dcomponents}. In this case, the function in eqn \eqref{3d-oft-f-no-4} achieves the best result via our visual validation. 

\begin{figure*}[!tp]
	\centering
	\begin{tabular}{cccc}
	    & Maximum & Mean  & Absolute deviation \vspace{0.08in}
	    \\
	    \rotatebox{90}{\hspace{0.15in} Line integral ${\cal R}[I]$} &
	    \includegraphics[width=0.27\textwidth]{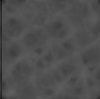} &
		\includegraphics[width=0.265\textwidth]{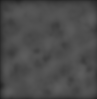} &
		\includegraphics[width=0.265\textwidth]{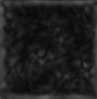} \\
		& (a) & (b) & (c) \\
	    \rotatebox{90}{\hspace{-0.05in} Alignment integral ${\cal G}[{\cal F}]$} &
	    \includegraphics[width=0.27\textwidth]{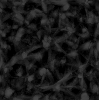} & 
		\includegraphics[width=0.265\textwidth]{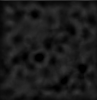} &
		\includegraphics[width=0.275\textwidth]{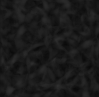} \\
		& (d) & (e) & (f) 	 
    \end{tabular}
\caption{Maximum, mean and absolute deviation of the line integral ${\cal R}[I]$ and alignment integral ${\cal G}[{\cal F}]$ on test image in Fig. \ref{origimages} (c). Columns from left to right respectively give the maximum, mean and absolute deviation of the line integral (first row) and alignment integral (second row). }
\label{3dmodules}
\end{figure*}

\begin{figure*}[!tp]
\centering
\begin{tabular}{ccc}
\includegraphics[width=0.27\textwidth]{./media/origplb.png} & 
\includegraphics[width=0.275\textwidth]{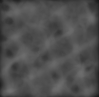} & 
\includegraphics[width=0.27\textwidth]{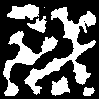} \\
(a) & (b) & (c)
\end{tabular}
\caption{Performance of the proposed orientation field transform on the image in Fig. \ref{origimages} (c).
(a): given image; (b): the result of the proposed orientation filed transform using $f$ defined in \eqref{3d-oft-f-1-3}; (c): the segmentation result obtained by hard thresholding of (b).}
\label{vplb}
\end{figure*}

\begin{figure}[!tp]	
\centering	
\begin{tabular}{c}
\includegraphics[width=0.383\textwidth]{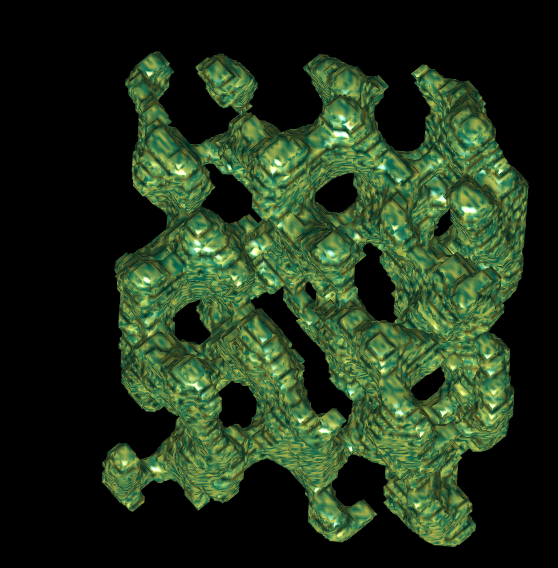} \\
(a) \\
\includegraphics[width=0.38\textwidth]{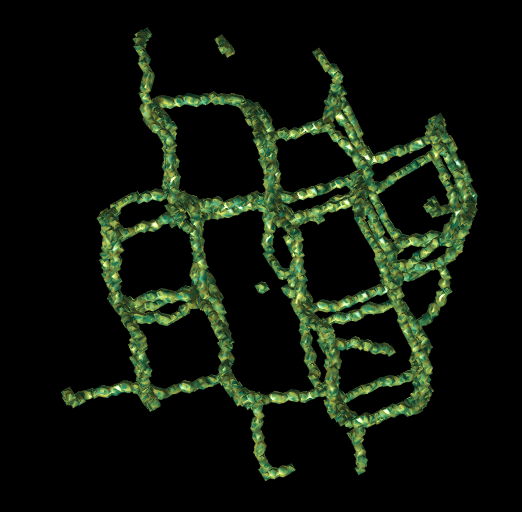} \\
(b) 
\end{tabular}
\caption{Performance of the proposed orientation field transform on the 3D image in Fig. \ref{origimages} (d).
(a): the result of the proposed orientation filed transform using $f$ defined in \eqref{3d-oft-f-1-3} (morphological closing is used for a better view); (b): medial axis transform of (a).}
\label{3dplb}
\end{figure}

\begin{figure*}[!tp]
\centering
\begin{tabular}{cccc}
& $\langle 1\ 0\ 0 \rangle$ & $\langle 1\ 1\ 1 \rangle$ & $\langle 1\ 1\ 0 \rangle$ \vspace{0.05in} \\
\includegraphics[width=0.23\textwidth]{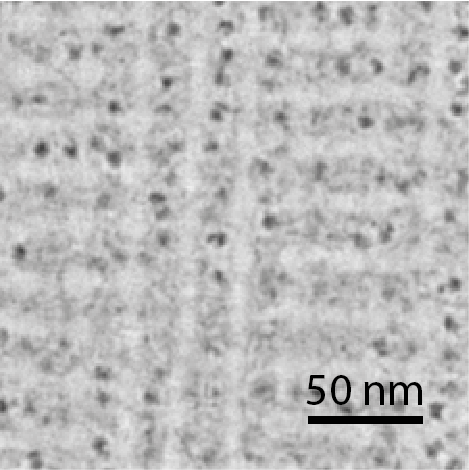} & 
\includegraphics[width=0.22\textwidth]{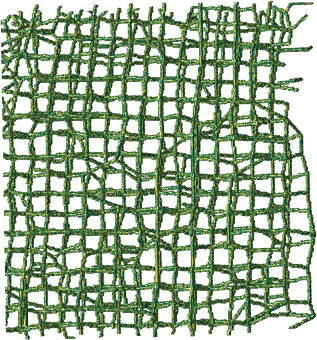} & 
\includegraphics[width=0.22\textwidth]{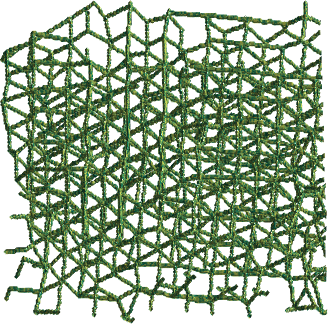} &
\includegraphics[width=0.22\textwidth]{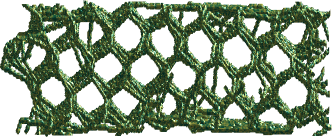} \\
(a) & (b) & (c) & (d) \\
 & 
\includegraphics[width=0.22\textwidth]{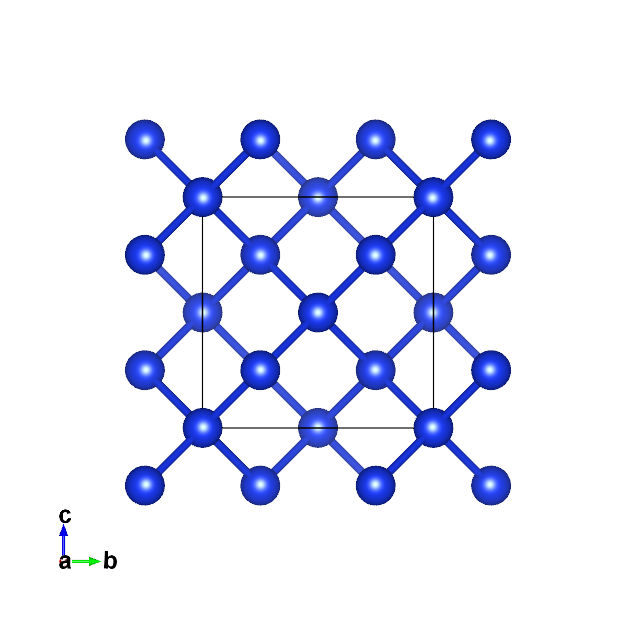} & 
\includegraphics[width=0.22\textwidth]{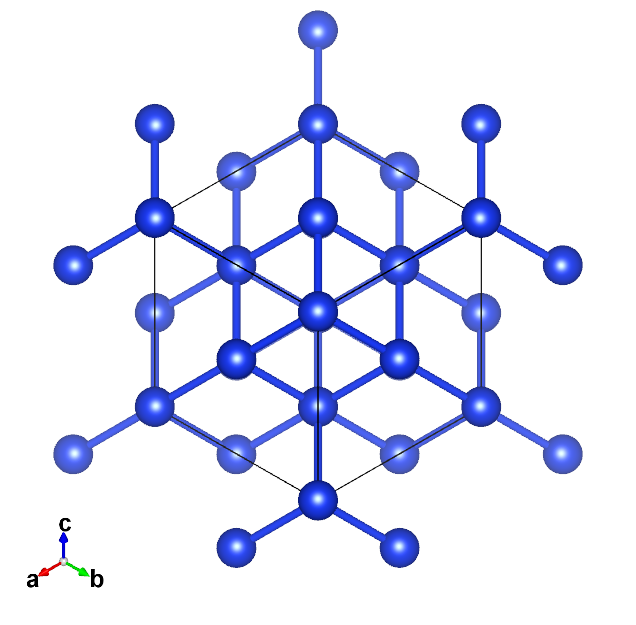} & 
\includegraphics[width=0.22\textwidth]{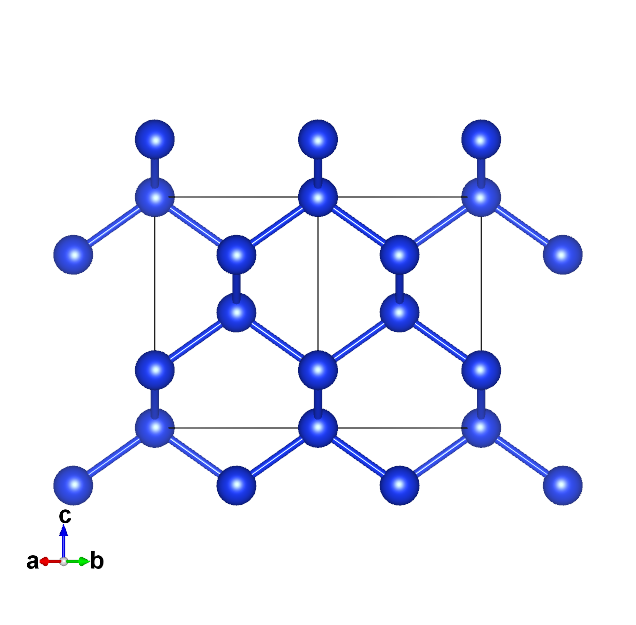}  \\
& (e) & (f) & (g)
\end{tabular}
\caption{Performance of the proposed orientation field transform on a greater cropped 3D region of the liquid crystal imaged with electron tomography. (a): given image; (b)--(d): skeletons of the segmentation result following the same method as in Fig. \ref{3dplb} with skeleton denoising viewed at $\langle 1\ 0\ 0 \rangle$, $\langle 1\ 1\ 1 \rangle$ and $\langle 1\ 1\ 0 \rangle$, respectively; (e)--(g): a cubic diamond lattice unit cell rendered with VESTA and viewed at $\langle 1\ 0\ 0 \rangle$, $\langle 1\ 1\ 1 \rangle$ and $\langle 1\ 1\ 0 \rangle$, respectively.}
\label{groundtruth}
\end{figure*}

\subsubsection{Real-world 3D image}
The proposed method is now tested on the real-world 3D image in Fig. \ref{origimages} (c)--(d). The curve detection in this image is extremely challenging since the curve information is even barely visually sensible. As mentioned previously, it is an image of a lyotropic liquid crystal, whose curves converge and diverge in different directions frequently and regularly. A vast number of curves meanders along different directions at close proximity and crams next to each other.

The performance of the measures of the maximum, mean and absolute deviation of the line integral ${\cal R}[I]$ and alignment integral ${\cal G}[{\cal F}]$ are presented in Fig. \ref{3dmodules}. It shows that the maximum and mean of the line integral perform better than other measures in enhancing the obscure curves. The close packing of the curves might have negated the need to exclusively remove structures with a clear orientation.
Therefore, it is wise to use the function $f$ in eqn \eqref{3d-oft-f-1-3} in the proposed transform for this test image. The enhanced curves and the subsequent segmentation result with hard thresholding are shown in Fig. \ref{vplb}, which indeed presents curve features that are imperceptible in the given image. 
The 3D view of the segmentation results across the entire volume of the testing image of the lyotropic liquid crystal in Fig. \ref{origimages}(c) is given in Fig. \ref{3dplb}. Note that as the values of the integrals decrease for the pixels at the periphery, each $x$-$y$ plane across the $z$-axis is linearly scaled to have the same median value before hard thresholding of the curve-enhanced result. The lyotropic crystal is triply periodic, so it is expected to be `seen through' over several layers of periodicity when viewed from several angles, with regular interruptions on the viewing plane. A medial axis transform (skeletonisation) is performed on the segmented image Fig. \ref{3dplb} (a) with a MATLAB function to better show the segmentation quality, see Fig. \ref{3dplb} (b). 

A greater cropped 3D region of the liquid crystal is shown
in Fig. \ref{groundtruth} (a), which is known to take on a diamond cubic symmetry. As a demonstration, along the viewing directions shown in Fig. \ref{groundtruth}, the lattice viewed at $\langle 1\ 0\ 0 \rangle$ should appear as tessellating squares (Fig. \ref{groundtruth} (e)), that viewed at $\langle 1\ 1\ 1 \rangle$ should appear as tessellating triangles (Fig. \ref{groundtruth} (f)), and that viewed at $\langle 1\ 1\ 0 \rangle$ should appear as tessellating hexagons (Fig. \ref{groundtruth} (g)). 
The diamond-cubic lattice is then compared against the result (Fig. \ref{groundtruth} (b)--(d)) obtained using the same method as in Fig. \ref{3dplb} with a skeleton denoising procedure (see Appendix). The results are in congruence with the diamond cubic lattice structure, proving that the proposed 3D orientation transform is sufficient for the curve enhancement and segmentation quality we were seeking.

\section{Conclusion}
\label{sec:oftsummary}
An orientation field-based 3D orientation field transform was proposed and experimented for the curve enhancement of a liquid crystal, with segmentation as a byproduct. That being said, the proposed 3D orientation field transform does enhance curves selectively and effectively, which can also work as a preliminary filter for mixing with other segmentation and denoising methods. Even though this is a top-down processing transform, it involves only a few computational steps, and hence would serve as an ideal candidate as a preliminary filter. In consequence, the combination of the maximum, mean and absolute deviation of line integrals and alignment integrals was found to be an effective 3D orientation field transform for extremely challenging synthetic and real-world images. Furthermore, the proposed 3D orientation field transform can naturally tackle any dimensions.

Critical future work may follow the investigation of the impact of the single parameter $\varepsilon$ (the length of the paths for the integral) on the performance of the proposed 3D orientation field transform, and the search of an optimal function $f$ used in the proposed 3D orientation field transform. Moreover, the pursuit of utilising the enhanced images by the proposed 3D orientation field transform in artificial intelligence techniques is also of great interest.

\section*{Appendix}

\subsection{High-pressure freezing, sample processing, and microscopy}
The samples were prepared as described by \cite{IEEEhowto:liang}. \emph{Arabidopsis} cotyledons were dissected and frozen by an HPM100 high-pressure freezer. The samples were then freeze-substituted on planchettes at $-80$ $^{\circ}$C for 24~h and after that slowly warmed up to room temperature for over 48~h. The warm samples were transferred to be embedded in 812 resin and polymerized in an oven at 65 $^{\circ}$C, which was then sliced to 200~nm thick sections with an ultramicrotome. In the end, the samples were examined with a 200~kV Tecnai F20 intermediate voltage electron microscope.

\subsection{Skeleton denoising used in Fig. \ref{groundtruth}}
The medial axis transform on noisy tubules may occasionally produce an artefact of separated nodes that should in fact be merged, especially on nodes that have a high number of connecting tubules (high degree). Moreover, it is the connectivity (topology) between the intersections (nodes) of the tubules that are of relevance to the lattice structure. Hence, the skeletonised tubules are straightened out and some of the nodes are merged for denoising, which is done with the following steps:
\begin{enumerate}
		\item Convert the skeleton into an undirected adjacency matrix with node coordinates using the algorithm {\tt Skel2Graph3D} developed by \cite{kollmannsberger2017small}.
		\item Average the coordinates of nodes that have a neighbouring distance lower than a selected threshold value (i.e., 5.4~nm).
		\item Repeat step 2 until the coordinates stop changing.
		\item Plot straight lines to connect back the nodes using the Bresenham's line algorithm with the coordinates data and the adjacency matrix. In particular, the undocumented MATLAB adaptation of the Bresenham's line algorithm {\tt iptui.intline} is modified to process the 3D data.
\end{enumerate}

%

\ifCLASSOPTIONcaptionsoff
  \newpage
\fi



\bibliographystyle{IEEEtran}
\bibliography{reference_3d-oft}
%

%








\end{document}